# From Data to Diagnosis: A Large, Comprehensive Bone Marrow Dataset and AI Methods for Childhood Leukemia Prediction

**Running title**

Bone Marrow Dataset & Methods for Childhood Leukemia

**Authors**

Henning Höfener[1*], Farina Kock[1], Martina Pontones[2], Tabita Ghete[2,3], David Pfrang[1], Nicholas Dickel[4], Meik Kunz[4], Daniela P Schacherer[1], David A Clunie[5], Andrey Fedorov[6], Max Westphal[1], Markus Metzler[2,3,7]

[1] Fraunhofer Institute for Digital Medicine MEVIS, Bremen, Germany

[2] Department of Pediatrics and Adolescent Medicine, University Hospital Erlangen, Erlangen, Germany

[3] Bavarian Cancer Research Center (BZKF), Erlangen, Germany

[4] Medical Informatics, Friedrich-Alexander University of Erlangen-Nürnberg, Erlangen, Germany

[5] PixelMed Publishing LLC, Bangor, PA, USA

[6] Department of Radiology, Brigham and Women's Hospital and Harvard Medical School, Boston, MA, USA

[7] Comprehensive Cancer Center Erlangen-EMN (CCC ER-EMN), Erlangen, Germany

[*] Corresponding author (henning.hoefener@mevis.fraunhofer.de)

# From Data to Diagnosis: A Large, Comprehensive Bone Marrow Dataset and AI Methods for Childhood Leukemia Prediction

## Abstract

Leukemia diagnosis primarily relies on manual microscopic analysis of bone marrow morphology supported by additional laboratory parameters, making it complex and time consuming. While artificial intelligence (AI) solutions have been proposed, most utilize private datasets and only cover parts of the diagnostic pipeline. Therefore, we present a large, high-quality, publicly available leukemia bone marrow dataset spanning the entire diagnostic process, from cell detection to diagnosis. Using this dataset, we further propose methods for cell detection, cell classification, and diagnosis prediction. The dataset comprises 246 pediatric patients with diagnostic, clinical and laboratory information, over 40 000 cells with bounding box annotations and more than 28 000 of these with high-quality class labels, making it the most comprehensive dataset publicly available. Evaluation of the AI models yielded an average precision of 0.96 for the cell detection, an area under the curve of 0.98, and an F1-score of 0.61 for the 33-class cell classification, and a mean F1-score of 0.90 for the diagnosis prediction using predicted cell counts. While the proposed approaches demonstrate their usefulness for AI-assisted diagnostics, the dataset will foster further research and development in the field, ultimately contributing to more precise diagnoses and improved patient outcomes.

# Introduction

Bone marrow (BM) morphology is the basis for the assessment of hematopoiesis and many systemic diseases. In cancer research, BM morphology can be used to analyze both physiological changes in organ function and pathological processes in neoplasia. These changes are reflected in quantitative and qualitative effects that can only partially be captured by the current analog approach. In clinical practice, BM collection, preparation, and analysis is still largely manual and dependent on the available equipment, the used procedures, and the experience of the medical staff. This variability limits the comparability of manually collected data and can lead to uncertainties in disease classification(1). The diagnosis of leukemia for both children and adults can typically be made based on the manual differential cell count (DCC) of approximately 100-200 nucleated BM cells in a bone marrow aspirate (BMA) smear. DCC is routinely performed by specially trained hematologists with years of experience, but the process can be slow, error-prone, and resource-intensive. Furthermore, the distinction of some therapeutically relevant subtypes requires not only BM morphology, but also additional clinical and laboratory parameters(2). Due to the complexity of the data, the diagnostic process for leukemia is challenging and is highly dependent on the experience of the clinicians involved.

Therefore, automation of these processing steps by artificial intelligence (AI) models has the potential to overcome the current limitations of BM analysis. Several approaches of AI-based detection and classification of BM cells, as well as diagnosis predictions have already been proposed. We have previously published a comprehensive review of current AI

approaches(3). The training of high-performing AI models and their evaluation require large and high-quality datasets. However, most datasets used in the literature are kept private. Only few adequately sized datasets have been published so far. For BM cell detection, Eckardt, Schmittmann, et al.(4) released a dataset containing ~5k cells in 62 regions of interest (ROI) images. For cell detection in peripheral blood, a number of smaller datasets ranging from ~1.5k to ~40k cells have been published(5–7). While more datasets exist for cell classification, only few incorporate more than 10k cells and more than 10 cell classes(4,8–10). Of these, Matek et al.'s dataset(9) gained the most traction and has been used for developing several classification approaches(11–17). Datasets aimed at leukemia diagnosis prediction that feature blood or BM image data are particularly scarce and focus only on adult leukemia patients(14,18,19).

While the fundamental principles of BM analysis remain consistent across both children and adults, the absence of datasets specifically targeting childhood leukemia is a significant limitation, given the reported variations in the distribution of specific leukemia types in childhood(20–22). Furthermore, all of the mentioned datasets are only suitable for parts of the processing steps listed above.

This work presents a large and comprehensive dataset for the analysis of BMA smears from 246 pediatric patients with leukemia. It encompasses over 40 000 cell annotations and high-quality, fine-grained class labels for more than 28 000 cells, obtained through a consensus approach from five hematology experts. It is the first dataset of this size to integrate three pivotal steps of the leukemia diagnosis process: cell detection, cell classification, and diagnosis prediction. In addition, this work introduces a suite of capable methods for these three processing steps, which have been trained and evaluated using this dataset.

# Methods

## Dataset

This study retrospectively selected a cohort of 246 patients who met the inclusion criteria. The criteria required patients to be diagnosed with acute myeloid leukemia (AML), chronic myeloid leukemia (CML), or acute lymphoblastic leukemia (ALL) and to be under 18 years old at the time of initial diagnosis. Additionally, we required a BMA smear stained using the Pappenheim staining method to be available at the initial diagnosis, prior to any treatment. Please refer to Supplement 1 for a comprehensive list of inclusion criteria, along with a detailed list of task-specific exclusion criteria. This research has been approved by the ethics committee of the Friedrich-Alexander-University Erlangen-Nuernberg (Application 22-148-Br). All participants have given written informed consent.

BMA smears were digitized using a 3DHistech Pannoramic MIDI II whole slide scanner with 40x objective lens (without immersion), resulting in a resolution of 0.11x0.11 µm/pixel. Two smears were digitized with a 3DHistech Pannoramic 250 Flash scanner. All images were initially saved as proprietary MRXS format files, but afterwards converted to standard DICOM format for sharing.

Square regions of interest (ROI) of 2048 pixels edge length with well-spread (individually separated) cells were annotated on most of the digitized BMA smear 40x images. To generate the cell detection dataset, for a subset of the ROIs, all leukocytes within it were delineated with bounding box annotations. For the classification dataset, single-cell images of 256x256 pixels were extracted from the center of each bounding box. In cases where a cell exceeded these dimensions, a square image was extracted containing the complete bounding box. Images larger than 4096 pixels were downsampled afterwards. To enhance the dataset, we further included some erythrocytes, thrombocytes, and rare cell types from various areas of the smears following the same procedure as outlined above.

To achieve high quality yet efficient cell class labeling of the single-cell images, we used a custom consensus approach, where each image is successively annotated by different observers until two requirements are met: An image must be annotated by at least two observers and the majority class must be selected in at least half of all annotations for that image. This consensus labelling approach was implemented as a multi-user web application that was developed for this purpose. The labeling view was particularly designed for high efficiency. Assigning a class label only requires a single click in the hierarchical, color-coded class tree. Figure 1 shows a screenshot of the labeling view of the annotation application. For the second observation of the cells, a validation mode was implemented. In this mode, cells that have been assigned to a particular cell class are displayed in a gallery view. This allows observers to check the gallery for cells that they consider to be incorrectly labeled and correct them. Although this procedure might introduce confirmation bias, it was incorporated to enhance the efficiency of the annotation process.

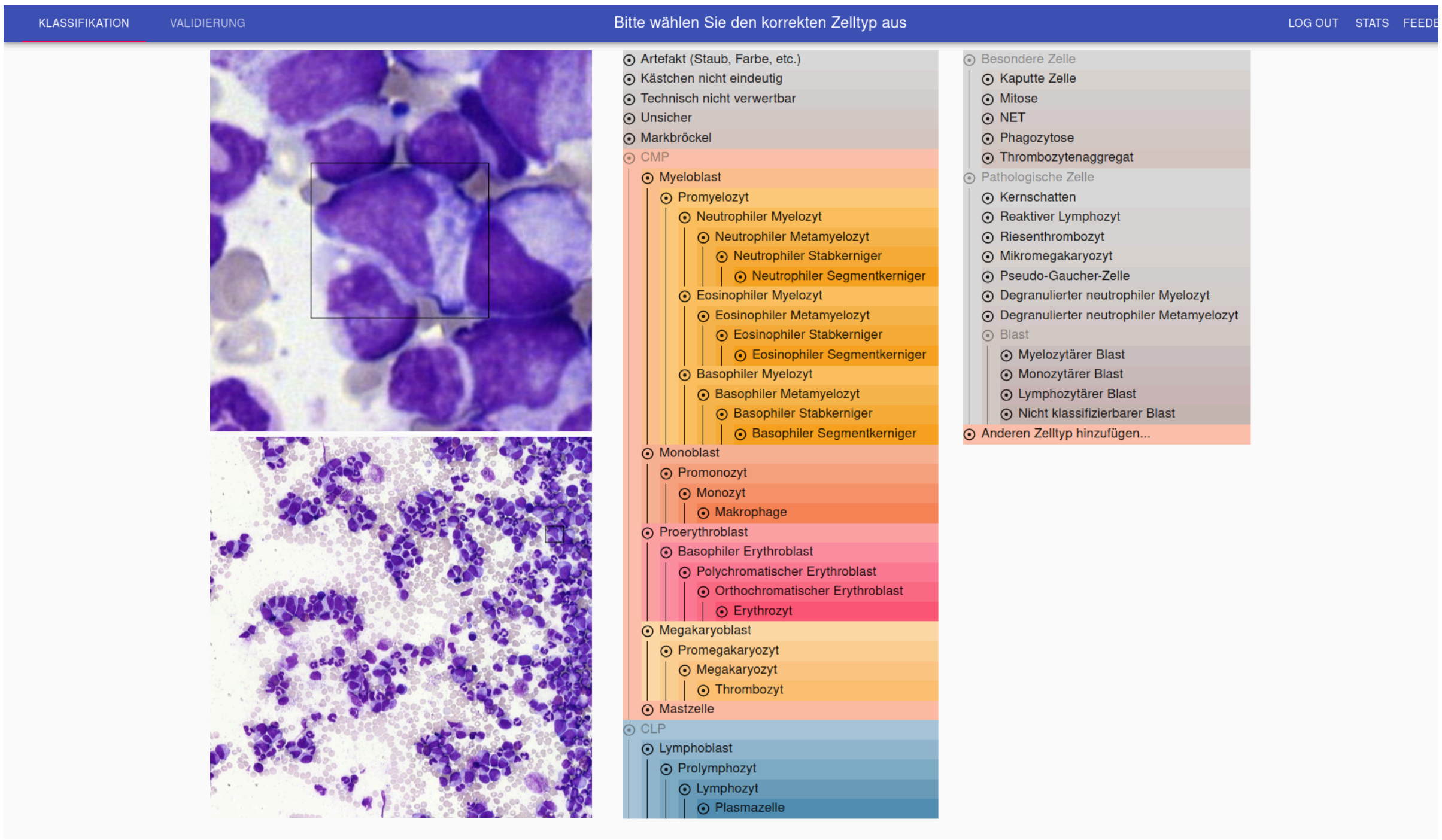

**Figure 1:** Web application for cell labeling. The screenshot shows the labeling view with the single-cell image and an overview image on the left, as well as a color-coded class tree on the right.

The annotation application was implemented to assign cells to one of >50 classes. For cell classification, we mapped it to 33 classes, as several classes had too few samples (Supplemental Table 1). The cells for which a consensus was reached comprise the cell classification dataset.

The ROIs, bounding boxes, and cell class annotations were stored in an internal proprietary format/database, and for sharing have been converted into a standard DICOM annotation format.

Clinical data was collected including leukemia type and subtype, gender and age (in intervals). Laboratory values also include the manual DCC as reported during clinical routine. Data was gathered from multiple different information systems, where not all requested parameters were necessarily available. Accordingly, the laboratory data contains a considerable number of missing values. Thus, a trade-off between clinical relevance and the completeness of the features had to be determined. This resulted in a total of 18 laboratory values (Supplemental Table 2).

The dataset was partitioned into a training, a validation, and a test set at patient level using 60%, 20%, 20% of patients, respectively. This process was executed in a two-step manner. Initially, each patient was allocated to one of the sets with the objective to achieve an equitable distribution of diagnoses across all sets. Subsequently, patients with the same diagnoses were swapped between sets, ensuring that the cell class distribution across all sets was as close as possible. This manual process was implemented to ensure that the sets exhibited similar distributions of both diagnoses and cell classes, thereby ensuring that they are suitable for all tasks of the analysis pipeline. The specific stratification of patients into the sets is documented in Supplemental Table 3.

## Automated Differential Cell Count

While some approaches integrate cell detection and classification into a single model, we opted to train two separate models for these tasks. This allows for individual and decoupled optimization and evaluation of each stage and improves comparability to other methods.

For the cell detection, we used two different state-of-the-art approaches being CenterNet(23) and Faster R-CNN (FRCNN)(24). The CenterNet uses an ImageNet-pretrained ResNet-50 backbone and outputs objects of a single class. It was trained for 50 epochs with Adam optimizer (lr=1e-4). The images were augmented using standard methods. The confidence threshold was optimized using the validation set. For prediction, local maxima within a distance of 10 pixels were considered. The FRCNN approach uses a COCO-pretrained ResNet-50 backbone and was used to output bounding box predictions only. It was trained for 250 epochs using stochastic gradient descent as the optimizer (lr=1e-3). The number of box detections per image was set to 544, representing the maximum number of bounding boxes per image based on the training and validation sets. The non-maximum suppression threshold for the prediction head was reduced to 0.3 to suppress the output of bounding boxes with a larger overlap. Since the ROIs only contain well-spread cells, there is minimal or no overlap of cells in the dataset. We applied random flip, rotation, resizing, color jittering, and noise addition for data augmentation. We measured precision, recall, and F1-score as well as the average precision (AP) to compare and evaluate the cell detection models.

To classify the cells into 33 classes, we trained an ImageNet-pretrained ResNet-50 where we replaced the final linear layer with a dropout layer (dropout rate 0.2) followed by a concatenation layer where the output of the dropout is fused with the normalized resolution value that was stored alongside the single-cell images. This was then fed into a multilayer perceptron with one hidden layer of size 32 and an output layer with 33 class outputs. The model was trained with cross entropy loss for 85 epochs using the AdamW optimizer (lr=3e-4) and a batch size of 32. Learning rate was reduced (x10) if no improvement in the validation median F1-score was observed for 10 epochs. No class balancing was applied and the best model was selected according to the median F1-score on the validation set. The images are augmented as described above. Larger images were then downsampled to 256x256 pixels, then a random crop of 244x244 pixels was extracted. The cell classification approaches were compared and evaluated using the median and mean of the per-class F1-scores. Additionally, we evaluate both top 1 and top 2 accuracy (micro average) as well as the area under the receiver operating characteristic curve (AUROC; macro-average, one-vs-rest).

## Diagnostic Models

For determining the leukemia type of patients, we trained machine learning models to distinguish between AML, ALL and CML. As input features for our models, we used both manual DCCs created by hematology experts as part of clinical routine and automated

DCCs predicted by our cell detection and cell classification model pipeline. To compare the predictive performance of the manual and the automated DCC, we trained separate models for these two feature sets. Furthermore, we used a set of 18 laboratory values that are typically assessed in the context of leukemia diagnosis, to train a stand-alone model, using only lab values, and to train models using both lab values and DCCs. This allows us to evaluate the performance and importance of the different feature sets for leukemia diagnostics. The F1-score was chosen as the primary metric for comparison and evaluation of diagnostic models.

For diagnosis prediction we used gradient boosting as it is considered state-of-the-art for tabular data(25). We used grid search and a repeated holdout split (20 iterations) stratified by leukemia subtype to tune the hyperparameters of the HistGradientBoostingClassifier. While far more efficient methods exist for hyperparameter optimization (e.g. Bayesian optimization), grid search was deemed feasible due to the small size of our dataset.

## Statistical Analysis

To quantify uncertainty in any performance estimates, we used the bias-corrected and accelerated (BCa) bootstrap approach with 1000 bootstrap iterations to calculate 95% confidence intervals. This nonparametric method is applicable to all metrics for the different tasks. Some of the main advantages of BCa bootstrap confidence intervals compared to competing approaches are that they are transformation-respecting and second order accurate in terms of their coverage probability(26). We employed a hierarchical version of the BCa bootstrap to take into account the hierarchical structure of our data with multiple cells per patient similar to the approach of Saravan et al.(27). Due to the exploratory nature of our analyses, we did not apply an adjustment for multiplicity.

# Results

## Dataset

Our cohort consists of 246 patients with AML, ALL, or CML diagnosis (Figure 2a). The distribution of patients (54.2% male and 45.8% female of all patients with known sex) was very similar to the national incidence of pediatric leukemia with a ratio of male to female of 54.4% and 45.6%, respectively(28). The dataset comprises diagnostic information (leukemia type and subtype), manual DCCs and 18 laboratory parameters (Supplemental Table 3) from clinical routine, as well as digitized BMA smears for all of the 246 patients. We only considered data that was available at the time of diagnosis but before any treatment has been performed.

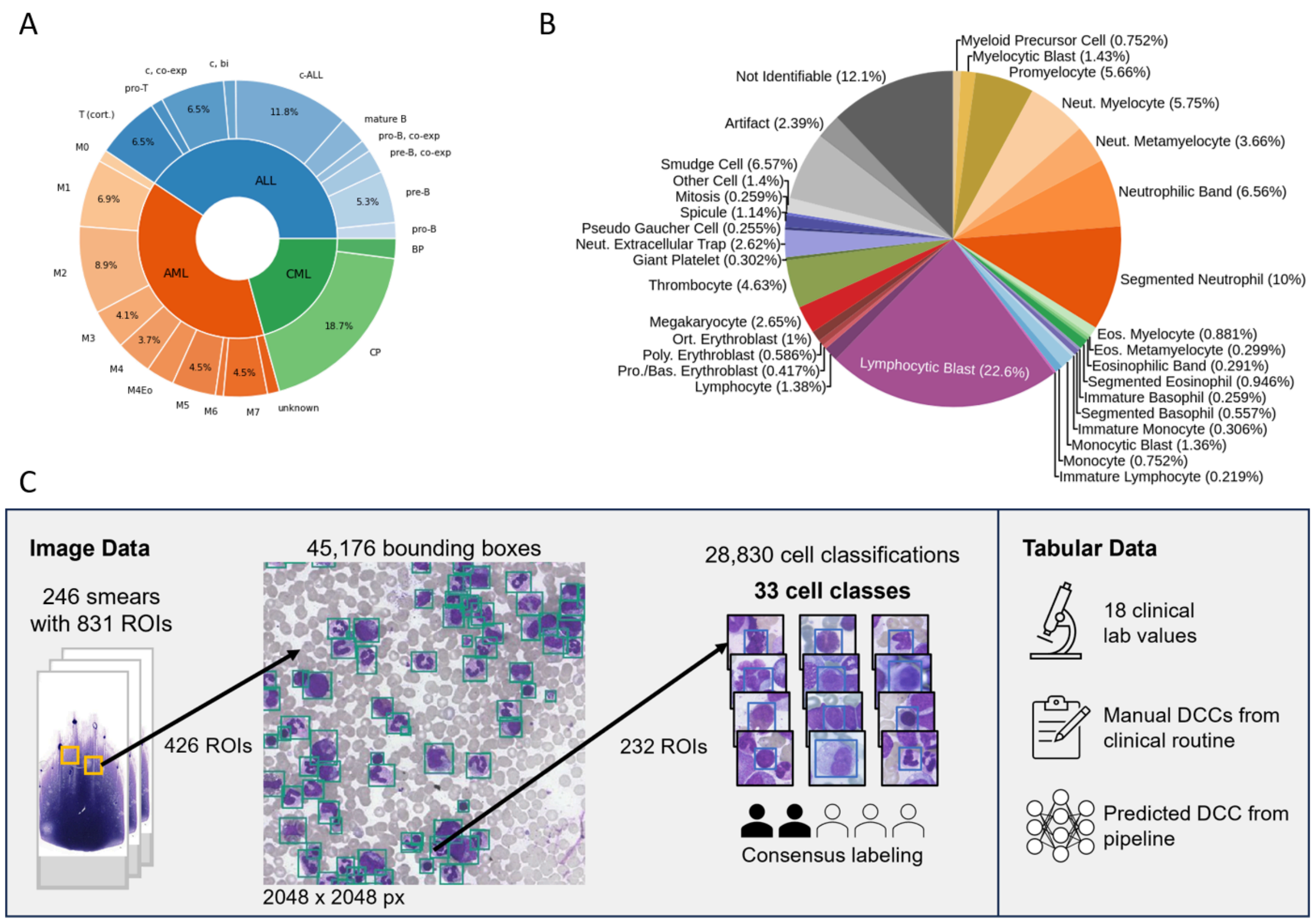

**Figure 2:** Overview of the structure and size of the dataset. A: Distribution of leukemia types and subtypes. B: Distribution of cell classes. C: Structure and size of the data and annotations.

A total of 831 ROIs were marked in the BMA smears. For 426 of them, all cells within the regions were exhaustively delineated using bounding boxes. For a subset of 232 of these ROIs, cell classes were annotated via a consensus labeling approach. This led to a total of 45,176 cells with bounding box annotations of which 28,830 cells also have class labels. Figure 2 visualizes the structure and size of the dataset.

To investigate the effect of our consensus labeling approach, we analyzed the annotations of the white blood cells performed by the five observers. For the majority of cells (87.8%) a consensus was already reached after two observations, which means that the first observation was confirmed by the second one. For another 8.4% of the cells consensus was reached after three observations. Here, the first two observations differed and the third observation confirmed one of the two previous observations. Performing more observations only led to a consensus in an additional 0.4% of the cells. The remaining 3.3% could not reach a consensus label. Of these, the cells belonging to the test set were labeled through a consensus meeting. To quantify the effect of our labeling approach in contrast to having only a single observer, we analyzed how many consensus labels differ from the first annotation, very likely constituting a correction of that annotation. This is the case for 47.2% of those cells for which the first two observations differed, being 5.6% of all cells.

As our labeling workflow was designed for high efficiency, we measured the time between a cell being shown and the selection of a label in the list. As the web application was often left open during breaks, leading to very long labeling times, we use median values to give a realistic estimate of the actual labeling time. We report a median of 3 seconds in the classification view and 1.2 seconds in the validation view, per cell.

# Automated Differential Cell Count

For cell detection, we evaluated both the CenterNet and the FRCNN. The results of the evaluation are listed in Table 1. The FRCNN clearly outperformed the CenterNet in all evaluated metrics.

| **Model** | **Precision** | **Recall** | **F1-score** | **Average Precision** |
|---|---|---|---|---|
| **CenterNet** | 0.890 | 0.904 | 0.897 | 0.919 |
| **FRCNN** | **0.967** | **0.945** | **0.956** | **0.958** |

**Table 1**: Precision, recall, F1-score, and average precision of the two cell detection approaches CenterNet and FRCNN for an intersection over union threshold of 0.5.

As shown in Table 2 and Figures 3 and 4, the cell classification shows a high prediction performance as visible in the confusion matrix. However, performance varies quite substantially between the classes, as can be observed in per-class F1-scores ranging from 0.000 (immature lymphocyte) to 0.945 (megakaryocyte). Performance is highly associated with the number of samples per class. 43% of classes with F1-scores below the median have less than 100 training samples, whereas this is only the case for 13% of the classes above the median. The high values of the top 2 accuracy indicate that the model is capable of learning the task but sometimes does not succeed in correctly ranking the top classes. The main reason is likely confusion between similar classes. This can also be observed in the confusion matrix, where we ordered the classes such that similar classes are close. Therefore, instead of a distinct diagonal, e.g. for eosinophils, the matrix builds up merely a block-like structure.

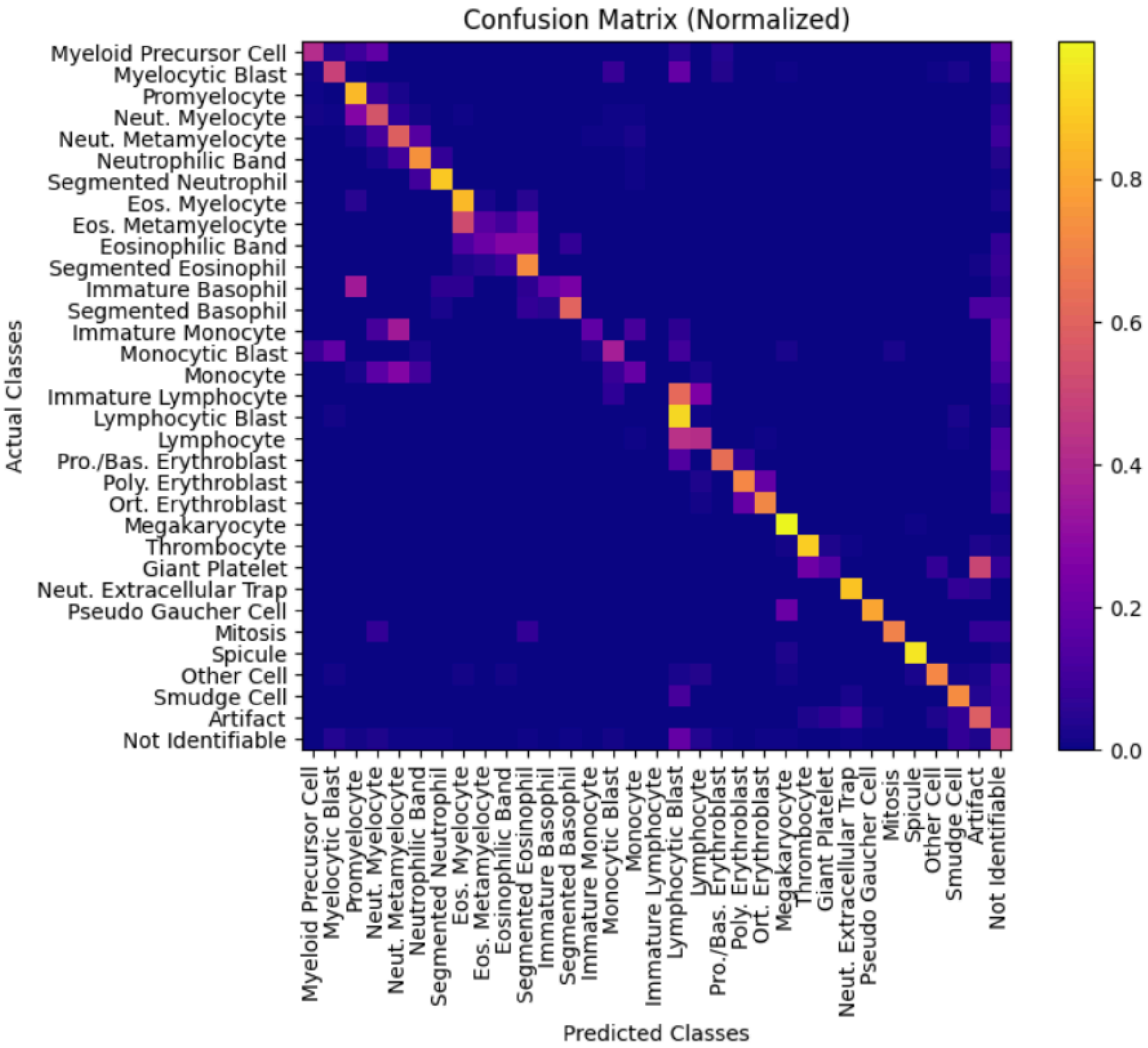

**Figure 3:** Confusion Matrix of the cell classification. Classes are ordered such that similar classes are close.

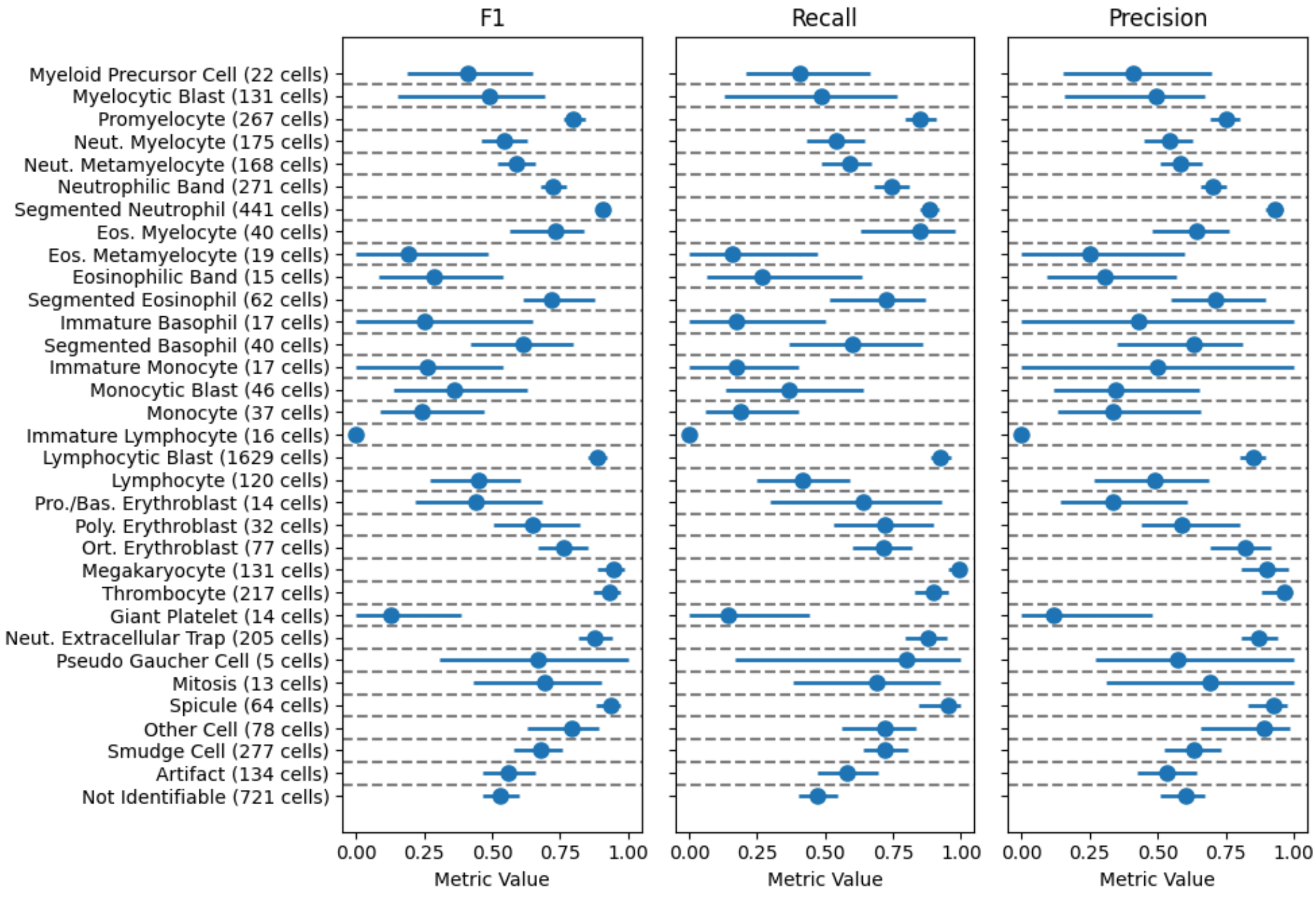

**Figure 4:** Visualization of the F1-scores, recall and precision values per cell class for the proposed cell classifier including 95% confidence intervals (BCa bootstrap, 1000 bootstrap iterations). The corresponding quantitative values are listed in supplemental Table 4.

As an external validation, we evaluated our approach using the dataset of Matek et al.(9). As this dataset only has one annotation per cell and the cell sizes are unknown, we modified our training procedure to not use this information (base model). To evaluate the effect of the proposed model vs. the base model, we also trained and evaluated the base model on our own dataset. Table 2 shows that our base model outperforms Matek. Furthermore, adding soft labels and cell masking appears to slightly improve the performance (Δ median F1 = 0.065; 95% CI: (-0.033, 0.124)). However, more test data would be required especially on rare cell classes to meaningfully measure this effect.

| **Model** | **Dataset** | **Median F1-score** | **Mean F1-score** | **Top 1 Accuracy** | **Top 2 Accuracy** | **AUROC** |
|---|---|---|---|---|---|---|
| **Matek** | Matek et al.(9) | 0.590 | 0.545 | N/A | N/A | N/A |
| **Base** | Matek et al.(9) | **0.749** **(0.731 - 0.767)** | **0.671** **(0.648 - 0.690)** | 0.873 (0.869 - 0.876) | 0.961 (0.959 - 0.963) | 0.990 (0.987 - 0.991) |
| **Base** | Ours | 0.550 (0.497-0.606) | 0.561 (0.534 - 0.591) | 0.729 (0.700 - 0.767) | 0.879 (0.858 - 0.899) | 0.982 (0.978 - 0.985) |
| **Proposed** | Ours | **0.615** **(0.557 - 0.680)** | **0.577** **(0.543 - 0.618)** | **0.748** **(0.721 - 0.782)** | **0.890** **(0.872 - 0.908)** | **0.981** **(0.976 - 0.985)** |

**Table 2**: Aggregated performance metrics including 95% confidence intervals (BCa bootstrap, 1000 bootstrap iterations) for cell classification models. The upper part of the table shows the model as proposed by Matek et al.(9) (no confidence intervals available) compared to our base model (the proposed model without soft labels and cell masks) trained and evaluated using Matek et al.'s dataset. The lower part of the table shows our base model compared to the proposed model trained and evaluated using our dataset.

## Diagnostic Models

We trained and evaluated three diagnostic models for leukemia prediction (target classes: ALL, AML, CML) trained with different sets of input features: laboratory values, DCCs from clinical routine, and predicted DCCs generated by our cell detection and classification models. To ensure a fair comparison, we only included patients for which all three kinds of input data are available, leading to 129 patients (64 ALL, 18 AML, 47 CML) in total. All models were evaluated on the test set (consisting of 32 patients) in terms of confusion matrices (Figure 5), as well as averaged performance measures (Table 3).

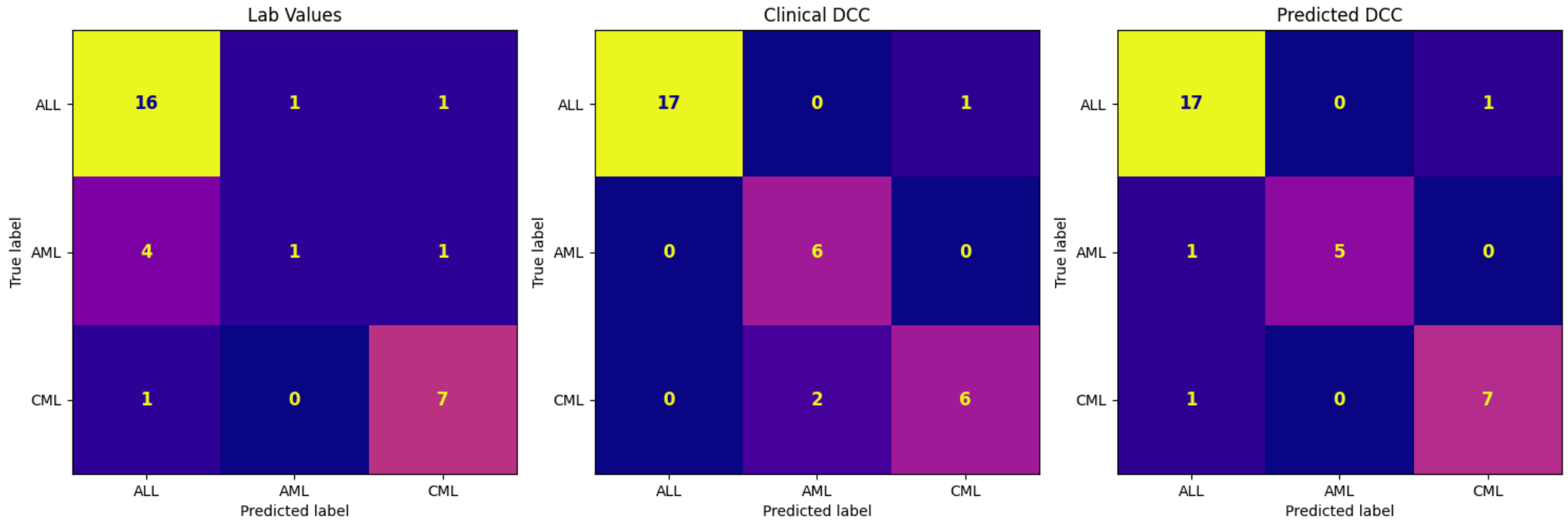

**Figure 5:** Confusion matrices on the test set for 3 diagnostic models using distinct sets of input features: laboratory parameters, DCCs from clinical routine, and predicted DCCs generated by our cell detection and classification models.

| **Model / Metric** | **F1-score ALL** | **F1-score AML** | **F1-score CML** | **Mean F1-score** |
|---|---|---|---|---|
| **Lab Values** | 0.821 (0.686 - 0.900) | 0.250 (0.000 - 0.667) | 0.824 (0.571 - 0.941) | 0.631 (0.496 - 0.834) |
| **Clinical DCC** | **0.971 (0.839-1.000)** | 0.857 (0.667 - 1.000) | 0.800 (0.462 - 0.941) | 0.876 (0.703 - 0.971) |
| **Predicted DCC** | 0.919 (0.789 - 0.973) | **0.901 (0.404 - 1.000)** | **0.875 (0.592 - 1.000)** | **0.901 (0.745 - 0.971)** |

**Table 3**: Mean and per-class F1 scores, including 95% confidence intervals (BCa bootstrap, 1000 bootstrap iterations) for uncertainty quantification.

While the lab values model has issues distinguishing ALL and AML patients, diagnostic models based on DCCs (both clinical and predicted) show better performance for ALL and AML (Table 3). The overall performance is similar for the predicted and clinical DCC models. Note that the high uncertainty in performance metrics is due to the small size of the test set.

We investigated whether using laboratory parameters and DCC data together yields better performance than using only DCCs. This was not the case on our dataset: exactly the same patients were misclassified after lab values have been added as further features, both for the clinical and the predicted DCCs.

# Discussion

AI-based hemato-morphology is an emerging field. A large number of approaches have been proposed in recent years. However, most approaches acquire and use internal datasets that are not shared. This substantially limits both reproducibility and the overall progress in the field, as researchers cannot build upon existing datasets. Although there are some publicly available datasets, none has emerged as a standard. Matek et al.'s dataset(9) is closest to being a standard benchmark, but their dataset has some limitations such as partitioning into training and test sets in a non patient-stratified manner, which allows overlap between slides of both sets and may compromise generalizability of trained models(29).

We created a large, high-quality, publicly shareable dataset for hematology that provides images and labels for detection and classification of digitized BMA smears, as well as patient-level clinical information and laboratory data for leukemia type prediction. While there are publicly available datasets for each of those tasks, ours is the only one of that size to combine them all. Comparing our dataset to other publicly available datasets, we can state that with >40k white blood cells, it is the largest dataset for cell detection. The second largest dataset comprises ~5k white blood cells(4). While there are larger datasets for cell classification with >92k cells(30) and >171k cells(9), our dataset is still large with >28k cells and is unique in that it employs a consensus labeling approach to ensure high-quality labels and is fine-grained to the extent that it distinguishes 33 different classes. In terms of diagnosis prediction, our dataset comprises 246 patients. A much larger cohort of 1251 AML patients with BMA smears was published for the prediction of NPM1 status(18). However, that dataset only contains a single leukemia type and only two laboratory values. In contrast, our dataset comprises three leukemia types with their subtypes and a set of 18 laboratory parameters.

But our dataset is not without limitations. Due to the considerable effort required to label a cell in a consensus meeting, only the cells from the test set were labeled in this manner. For the validation and training sets, these cells are designated as "no consensus found", which may impede the models' capacity to classify challenging samples. Also, as it is the case for virtually any dataset, a larger number of samples would have been beneficial. For both cell classes and leukemia subtypes, a substantial imbalance is observed, impeding the reasonable development of subtype classification models. While these limitations exist, our dataset nonetheless appears to be of exceptional value to the community.

Both steps of the automated DCC generation show high performance on the held-out test set, which was strictly kept unavailable for the development and optimization of the methods. The CenterNet and FRCNN approaches yield good detection results. The FRCNN's detection quality is comparable to that of the approach by Eckardt, Schmittmann et al.(4) with AP @ IoU<0.5 of 0.96 vs 0.97, confirming that the FRCNN approach is well suited for detecting leukocytes in BMA smears. While the FRCNN approach demonstrates superior performance metrics in comparison to CenterNet, the CenterNet remains a well-suited alternative in resource- and time-constrained environments.

The proposed cell classification approach shows high performance in the aggregated metrics. This is especially true given the very fine-grained classification task with 33 distinct classes. The approach also outperforms Matek et al.'s(9) classification model. However, difficulties for classes that share similar morphologies could be observed, as well as for classes with small sample sizes. The high top 2 accuracy shows that the model is capable of consistently ranking the correct class among its top predictions even if not always as the top choice.

Our diagnostic models are capable of distinguishing between ALL, AML, and CML, achieving a mean F1-score of 0.88 with clinical DCCs and 0.90 with predicted DCCs generated by our automated cell detection and classification pipeline. These results highlight the clinical potential of our DCC pipeline. The proposed set of 18 laboratory values alone does not effectively differentiate ALL from AML, nor does it provide additional value when combined with DCC data. This underscores the significance of DCC data as the most valuable feature for leukemia diagnostics, especially in the absence of genetic analysis, which is typically unavailable at the initial diagnosis stage.

The limited size of our dataset rendered the prediction of leukemia subtypes unfeasible. We distinguish 21 subtypes in our dataset (10 for ALL, 9 for AML, and 2 for CML), yet our reduced cohort of 129 patients used for training and evaluating diagnostic models lacked representation for some subtypes. Nonetheless, accurate diagnosis at the subtype level is crucial in clinical practice, for example in the distinguishing of acute promyelocytic leukemia, as treatment recommendations are contingent upon the specific subtype(31).

In this paper, we have introduced a large, high-quality pediatric leukemia bone marrow dataset along with methods for cell detection, classification, and diagnosis prediction. The dataset enables the development and evaluation of methods for the entire processing pipeline from cell detection through diagnosis prediction. To the best of our knowledge, it is the most comprehensive dataset for leukemia bone marrow analysis currently available.

Both proposed cell detection methods exhibit sufficient capability for real-world applications. The proposed cell classification approach yields high performance and outperforms Matek et al.'s(9) approach. Diagnostic models trained on the output of our cell classification pipeline are capable of distinguishing between ALL, AML and CML patients. This is a promising step in the direction of fully automated decision support in leukemia diagnostics.

Further extension of the dataset, particularly in regard to rare cell classes, is likely to enhance the automated DCC generation models. An expansion in leukemia types and subtypes would facilitate the development of diagnostic models for additional clinically relevant diagnostic classifications.

# Data sharing statement

The dataset introduced in this work will be made publicly available through the National Cancer Institute Imaging Data Commons(32), with the dataset managed through Zenodo and available at https://zenodo.org/records/15490664, where an exemplar subset of the data is already published(33).

# Acknowledgements

The authors thank Stefanie Barnickel, Nathalie Dollmann, Tatjana Flamann, Meinolf Suttorp, and Perdita Weller for the labelling of the cells.

The authors thank the following institutions for supplying BMA smears: University Hospital Augsburg (Univ.-Prof. Dr. Dr. med. Michael Frühwald), Charité Berlin - ALL-REZ BFM Study Group (PD Dr. med. Arend von Stackelberg), University Hospital at the Technical University Dresden (Prof. Dr. med. Meinolf Suttorp), University Hospital Essen - AML-BFM Study

Group (Prof. Dr. Dirk Reinhardt), Technical University of Munich (Prof. Dr. med. Irene Teichert-von Lüttichau), University Hospital Würzburg (Prof. Dr. med. Matthias Eyrich).

This study was supported by a grant from the German Federal Ministry of Education and Research (FKZ: 031L0262A; BMDeep)

Preparation of the Dataset for publication was partly supported by Federal funds from the National Cancer Institute, National Institutes of Health (Task Order No. HHSN26110071 under Contract HHSN261201500003I).

# Authorship Contributions

H.H., M.K., M.W., and M.M. designed the study; T.G. and M.P. acquired data and BMA smears and performed scanning; F.K., D.P., and N.D. curated patient data; H.H., F.K., and D.P. developed the AI methods; D.P. and M.W. performed statistical analysis; D.P.S., D.A.C., and A.F. performed integration of the dataset into IDC.

# Disclosure of Conflicts of Interest

The authors declare no conflicts of interest.

# Supplement

## Supplement 1: Inclusion and Exclusion criteria

Global Inclusion Criteria for the Research Project

- Children and adolescents up to the age of 18 at the time of initial diagnosis
- Limited to acute and chronic leukemia, specifically AML, ALL, CML
- Availability of the bone marrow smear at the time of initial diagnosis
- Uniform staining of the bone marrow smear according to Pappenheim
- The bone marrow smear must not contain any anticoagulants
- Intact bone marrow smears with airtight coverslip
- Availability of all information on the slide (relevant only for patients from University Hospital Erlangen)
    - Smear date
    - Name of the patient
    - BM identification (to distinguish from peripheral blood smear)
- Sufficiently good image quality (judged by hematology experts)
- The time period of retrospective inclusion varies per leukemia type and data source:
    - Patients from University Hospital Erlangen:
        - CML: 2013-2022
        - AML: 2011-2020
        - ALL: 2013-2021
    - Patients from other clinics:
        - No specific time restriction. Cases for specific rare ALL- and AML-subtypes were requested. Collected cases range from 2005 to 2022.

Task-Specific Exclusion Criteria and Patient Numbers

- **Cell Detection**
    - Included: 211 / 246 patients
    - Excluded: 35 / 246 patients. Reasons:
        - No cell boxes annotated: 35 patients

- **Cell Classification**
    - Included: 114 / 246 patients
    - Excluded: 132 / 246 patients. Reasons:
        - No cell classes annotated: 132 patients
- **Diagnostic Models**
    - Included: 129 / 246 patients
    - Excluded: 117 / 246 patients. Exclusion criteria (some patients fulfill more than one criterion):
        - Found out later that patient has already been under treatment at time of initial smear: 1 patient
        - No lab values available (meaning none at all; as soon as at least 1 lab value is available, this exclusion criterion does not apply. For some patients, only 4 / 18 lab values are available.): 116 patients
        - No clinical DCC available: 67 patients

# Supplemental Tables

| **Original Cell Type** | **Mapped Cell Type** |
|---|---|
| Myeloid Precursor Cell | Myeloid Precursor Cell |
| Myelocytic Blast | Myelocytic Blast |
| Promyelocyte | Promyelocyte |
| Neutrophilic Myelocyte | Neutrophilic Myelocyte |
| Degranulated Neutrophilic Myelocyte | |
| Neutrophilic Metamyelocyte | Neutrophilic Metamyelocyte |
| Degranulated Neutrophilic Metamyelocyte | |
| Neutrophilic Band | Neutrophilic Band |
| Segmented Neutrophil | Segmented Neutrophil |
| Eosinophilic Myelocyte | Eosinophilic Myelocyte |
| Eosinophilic Metamyelocyte | Eosinophilic Metamyelocyte |
| Eosinophilic Band | Eosinophilic Band |
| Segmented Eosinophil | Segmented Eosinophil |
| Basophilic Metamyelocyte | Immature Basophil |
| Basophilic Band | |
| Basophilic Myelocyte | |
| Segmented Basophil | Segmented Basophil |
| Promonocyte | Immature Monocyte |
| Immature Monocyte | |
| Monocytic Blast | Monocytic Blast |
| Monocyte | Monocyte |
| Prolymphocyte | Immature Lymphocyte |
| Lymphoid Precursor Cell | |
| Lymphocytic Blast | Lymphocytic Blast |
| Lymphocyte | Lymphocyte |
| Proerythroblast | Proerythroblast Or Basophilic Erythroblast |
| Basophilic Erythroblast | |
| Polychromatic Erythroblast | Polychromatic Erythroblast |
| Orthochromatic Erythroblast | Orthochromatic Erythroblast |
| Megakaryocyte | Megakaryocyte |
| Thrombocyte | Thrombocyte |
| Thrombocyte Aggregate | |
| Giant Platelet | Giant Platelet |
| Neutrophil Extracellular Trap | Neutrophil Extracellular Trap |
| Pseudo Gaucher Cell | Pseudo Gaucher Cell |
| Mitosis | Mitosis |
| Spicule | Spicule |
| Erythrocyte | Other Cell |
| Plasma Cell | |
| Macrophage | |
| Lymphoidocyte | |
| Phagocytosis | |
| Micromegakaryocyte | |
| Promegakaryocyte | |
| Smudge Cell | Smudge Cell |
| Artifact | Artifact |
| Damaged Cell | Not Identifiable |
| Technically Unfit | |
| Unknown Blast | |

**Supplemental Table 1**: Mapping of the original cell class labels as created using the annotation application to the resulting 33 cell classes used for cell classification.

| | **COMPONENT** | **DISPLAY_NAME** |
|---|---|---|
| **LOINC** | | |
| 14804-9 | Lactate dehydrogenase | LDH Lactate to pyruvate reaction |
| 1743-4 | Alanine aminotransferase | ALT With P-5'-P [Catalytic activity/Vol] |
| 20570-8 | Hematocrit | Hematocrit (Bld) [Volume fraction] |
| 2160-0 | Creatinine | Creatinine [Mass/Vol] |
| 28539-5 | Erythrocyte mean corpuscular hemoglobin | MCH (RBC) [Entitic mass] |
| 28540-3 | Erythrocyte mean corpuscular hemoglobin concentration | MCHC (RBC) [Mass/Vol] |
| 30239-8 | Aspartate aminotransferase | AST With P-5'-P [Catalytic activity/Vol] |
| 30385-9 | Erythrocyte distribution width | Erythrocyte distribution width (RBC) [Ratio] |
| 30428-7 | Erythrocyte mean corpuscular volume | MCV (RBC) [Entitic vol] |
| 3084-1 | Urate | Urate [Mass/Vol] |
| 6690-2 | Leukocytes | WBC Auto (Bld) [#/Vol] |
| 714-6 | Eosinophils/100 leukocytes | Eosinophils/100 WBC Manual cnt (Bld) |
| 718-7 | Hemoglobin | Hemoglobin (Bld) [Mass/Vol] |
| 737-7 | Lymphocytes/100 leukocytes | Lymphocytes/100 WBC Manual cnt (Bld) |
| 744-3 | Monocytes/100 leukocytes | Monocytes/100 WBC Manual cnt (Bld) |
| 769-0 | Neutrophils.segmented/100 leukocytes | Segmented neutrophils/100 WBC Manual cnt (Bld) |
| 777-3 | Platelets | Platelets Auto (Bld) [#/Vol] |
| 789-8 | Erythrocytes | RBC Auto (Bld) [#/Vol] |

**Supplemental Table 2**: List of laboratory values in the dataset as LOINC codes with their respective descriptions.

| **patient id** | **age** | **gender** | **leukemia_type** | **leukemia_subtype** | **split** | **exclusion_reason_cell_classification** | **exclusion_reason_cell_detection** | **exclusion_reason_diagnostic_models** |
|---|---|---|---|---|---|---|---|---|
| E2EF524FBF3D9FE611D5A8E90FEFDC9C | [4, 6[ | f | ALL | pro-B-ALL | train | | | |
| 02E74F10E0327AD868D138F2B4FDD6F0 | [16, 19[ | m | CML | Chronic Phase | train | | | No lab values available. |
| B53B3A3D6AB90CE0268229151C9BDE11 | [8, 11[ | m | ALL | c-ALL | train | No cell classes annotated for this patient. | | |
| C82561EC215A6E31807CEEDF3B3BD25E | [11, 14[ | f | CML | Chronic Phase | train | | | |
| F4B9EC30AD9F68F89B29639786CB62EF | [8, 11[ | f | ALL | pre-B-ALL | train | | | |
| A3F390D88E4C41F2747BFA2F1B5F87DB | [0, 3[ | f | ALL | intermediate cortical T-ALL | train | | | |
| 68D30A9594728BC39AA24BE94B319D21 | [4, 6[ | m | ALL | c-ALL | test | | | |
| 3416A75F4CEA9109507CACD8E2F2AEFC | [16, 19[ | m | CML | Chronic Phase | validation | | | |
| 4C56FF4CE4AAF9573AA5DFF913DF997A | [11, 14[ | m | AML | AML M2 | test | No cell classes annotated for this patient. | | |
| A1D0C6E83F027327D8461063F4AC58A6 | [8, 11[ | m | CML | Chronic Phase | validation | | | |
| 34173CB38F07F89DDBEBC2AC9128303F | [11, 14[ | m | CML | Chronic Phase | test | | | |
| 32BB90E8976AAB5298D5DA10FE66F21D | [4, 6[ | f | ALL | c-ALL | train | | | |
| 3295C76ACBF4CAAED33C36B1B5FC2CB1 | [4, 6[ | m | ALL | c-ALL | validation | No cell classes annotated for this patient. | | |
| 182BE0C5CDCD5072BB1864CDEE4D3D6E | [11, 14[ | f | CML | Chronic Phase | test | | | |
| DA4FB5C6E93E74D3DF8527599FA62642 | [4, 6[ | m | ALL | c-ALL, co-expression | train | | | |
| B6D767D2F8ED5D21A44B0E5886680CB9 | [6, 8[ | f | CML | Chronic Phase | train | | | |
| 02522A2B2726FB0A03BB19F2D8D9524D | [0, 3[ | m | AML | AML M5 | test | | | |
| A97DA629B098B75C294DFFDC3E463904 | [3, 4[ | f | ALL | c-ALL, co-expression | train | No cell classes annotated for this patient. | | |
| EC5DECCA5ED3D6B8079E2E7E7BACC9F2 | [6, 8[ | m | AML | AML M2 | validation | | | |
| 19CA14E7EA6328A42E0EB13D585E4C22 | [6, 8[ | m | CML | Chronic Phase | train | | | |
| 7CBBC409EC990F19C78C75BD1E06F215 | [6, 8[ | f | ALL | pre-B-ALL | train | | | |
| E4DA3B7FBBCE2345D7772B0674A318D5 | [16, 19[ | m | CML | Chronic Phase | train | | | |
| F0935E4CD5920AA6C7C996A5EE53A70F | [4, 6[ | m | ALL | intermediate cortical T-ALL | train | No cell classes annotated for this patient. | | Patient has already been under treatment at the time of the smear. |
| 38B3EFF8BAF56627478EC76A704E9B52 | [0, 3[ | m | ALL | pre-B-ALL | train | | | |
| AAB3238922BCC25A6F606EB525FFDC56 | [16, 19[ | m | CML | Chronic Phase | validation | | | |
| A684ECEEE76FC522773286A895BC8436 | [6, 8[ | m | ALL | intermediate cortical T-ALL | train | No cell classes annotated for this patient. | | |
| EB160DE1DE89D9058FCB0B968DBBBD68 | [14, 16[ | m | ALL | c-ALL | train | | | |
| 1F0E3DAD99908345F7439F8FFABDFFC4 | [6, 8[ | f | CML | Chronic Phase | train | | | |
| ED3D2C21991E3BEF5E069713AF9FA6CA | [14, 16[ | f | ALL | c-ALL | train | | | |
| 03AFDBD66E7929B125F8597834FA83A4 | [0, 3[ | m | ALL | pro-B-ALL | validation | No cell classes annotated for this patient. | | |
| C8FFE9A587B126F152ED3D89A146B445 | [8, 11[ | m | AML | AML M4 | test | | | |
| A87FF679A2F3E71D9181A67B7542122C | [14, 16[ | m | CML | Chronic Phase | test | | | |
| 6364D3F0F495B6AB9DCF8D3B5C6E0B01 | [8, 11[ | f | CML | Chronic Phase | train | | | |
| 42A0E188F5033BC65BF8D78622277C4E | [0, 3[ | m | AML | AML M7 | test | No cell classes annotated for this patient. | | |
| C51CE410C124A10E0DB5E4B97FC2AF39 | [16, 19[ | m | CML | Chronic Phase | train | | | |
| C9F0F895FB98AB9159F51FD0297E236D | [14, 16[ | m | CML | Chronic Phase | train | | | |
| D2DDEA18F00665CE8623E36BD4E3C7C5 | [3, 4[ | m | ALL | pre-B-ALL | train | | | |
| 54229ABFCFA5649E7003B83DD4755294 | [11, 14[ | m | ALL | c-ALL | train | | | |
| C20AD4D76FE97759AA27A0C99BFF6710 | [16, 19[ | m | CML | Chronic Phase | train | | | |
| 9BF31C7FF062936A96D3C8BD1F8F2FF3 | [14, 16[ | f | CML | Chronic Phase | test | | | No lab values available. |
| 1C383CD30B7C298AB50293ADFECB7B18 | [16, 19[ | m | CML | Chronic Phase | train | | | |
| C81E728D9D4C2F636F067F89CC14862C | [16, 19[ | m | CML | Chronic Phase | validation | | | |
| 98DCE83DA57B0395E163467C9DAE521B | [4, 6[ | f | ALL | c-ALL | validation | | | |
| 76DC611D6EBAAFC66CC0879C71B5DB5C | [8, 11[ | f | AML | AML M4 | train | | | |
| 3C59DC048E8850243BE8079A5C74D079 | [14, 16[ | m | CML | Chronic Phase | test | | | |
| 37693CFC748049E45D87B8C7D8B9AACD | [8, 11[ | m | CML | Chronic Phase | train | | | |
| ECCBC87E4B5CE2FE28308FD9F2A7BAF3 | [8, 11[ | f | CML | Chronic Phase | train | | | |
| 73278A4A86960EEB576A8FD4C9EC6997 | [11, 14[ | f | ALL | c-ALL, co-expression | validation | | | |
| D82C8D1619AD8176D665453CFB2E55F0 | [6, 8[ | f | ALL | pre-B-ALL | validation | No cell classes annotated for this patient. | | |
| 698D51A19D8A121CE581499D7B701668 | [0, 3[ | m | ALL | pre-B-ALL | train | | | |
| 5F93F983524DEF3DCA464469D2CF9F3E | [3, 4[ | f | ALL | c-ALL, co-expression | test | | | |
| 7F1DE29E6DA19D22B51C68001E7E0E54 | [16, 19[ | m | AML | AML M4 | validation | | | |
| 1679091C5A880FAF6FB5E6087EB1B2DC | [14, 16[ | m | CML | Chronic Phase | train | | | |
| AC627AB1CCBDB62EC96E702F07F6425B | [4, 6[ | f | ALL | c-ALL | train | | | |
| A5771BCE93E200C36F7CD9DFD0E5DEAA | [8, 11[ | f | CML | Chronic Phase | train | | | No lab values available. |
| 6F4922F45568161A8CDF4AD2299F6D23 | [4, 6[ | f | CML | Blast Phase | train | | | |
| 92CC227532D17E56E07902B254DFAD10 | [16, 19[ | m | ALL | c-ALL | train | | | |
| 28DD2C7955CE926456240B2FF0100BDE | [4, 6[ | f | ALL | c-ALL | validation | No cell classes annotated for this patient. | | |
| 66F041E16A60928B05A7E228A89C3799 | [0, 3[ | f | ALL | c-ALL | validation | No cell classes annotated for this patient. | | |
| 65B9EEA6E1CC6BB9F0CD2A47751A186F | [16, 19[ | f | ALL | c-ALL | train | No cell classes annotated for this patient. | | |
| 44F683A84163B3523AFE57C2E008BC8C | [6, 8[ | m | ALL | intermediate cortical T-ALL | test | | | |
| 33E75FF09DD601BBE69F351039152189 | [8, 11[ | f | CML | Chronic Phase | train | | | |
| 0F28B5D49B3020AFEECD95B4009ADF4C | [16, 19[ | m | AML | AML M3 | train | No cell classes annotated for this patient. | | |
| A5BFC9E07964F8DDDEB95FC584CD965D | [8, 11[ | m | CML | Chronic Phase | train | | | |
| F033AB37C30201F73F142449D037028D | [4, 6[ | m | ALL | c-ALL | train | | | |
| 07E1CD7DCA89A1678042477183B7AC3F | [4, 6[ | f | ALL | intermediate cortical T-ALL | test | No cell classes annotated for this patient. | | |
| 069059B7EF840F0C74A814EC9237B6EC | [0, 3[ | m | AML | AML M5 | validation | | | |
| 98F13708210194C475687BE6106A3B84 | [11, 14[ | m | CML | Chronic Phase | train | | | |
| E369853DF766FA44E1ED0FF613F563BD | [16, 19[ | f | CML | Chronic Phase | train | | | |
| 202CB962AC59075B964B07152D234B70 | [8, 11[ | f | AML | AML M2 | train | | | |
| 17E62166FC8586DFA4D1BC0E1742C08B | [14, 16[ | m | CML | Chronic Phase | train | | | |
| C4CA4238A0B923820DCC509A6F75849B | [11, 14[ | m | CML | Chronic Phase | train | | | |
| E2C420D928D4BF8CE0FF2EC19B371514 | [14, 16[ | m | ALL | pro-T-ALL | test | | | |
| 8E296A067A37563370DED05F5A3BF3EC | [16, 19[ | f | CML | Chronic Phase | test | | | |
| D1FE173D08E959397ADF34B1D77E88D7 | [8, 11[ | f | ALL | pre-B-ALL | validation | | | |
| 642E92EFB79421734881B53E1E1B18B6 | [4, 6[ | m | CML | Chronic Phase | train | | | |
| 9B8619251A19057CFF70779273E95AA6 | [14, 16[ | m | AML | AML M5 | train | | | |
| EA5D2F1C4608232E07D3AA3D998E5135 | [0, 3[ | m | ALL | c-ALL | validation | No cell classes annotated for this patient. | | |
| 70EFDF2EC9B086079795C442636B55FB | [8, 11[ | f | CML | Chronic Phase | train | | | |
| 8EB6D8747175919F27472673A32B3ECA | [8, 11[ | m | CML | Blast Phase | test | No cell classes annotated for this patient. | | |
| E00DA03B685A0DD18FB6A08AF0923DE0 | [11, 14[ | m | AML | AML M1 | train | No cell classes annotated for this patient. | | |
| D3D9446802A44259755D38E6D163E820 | [14, 16[ | m | CML | Chronic Phase | validation | | | |
| 14BFA6BB14875E45BBA028A21ED38046 | [0, 3[ | f | ALL | c-ALL | train | | | |
| D9D4F495E875A2E075A1A4A6E1B9770F | [16, 19[ | m | CML | Chronic Phase | train | | | |
| 7647966B7343C29048673252E490F736 | [11, 14[ | f | ALL | pre-B-ALL | train | | | |
| 65DED5353C5EE48D0B7D48C591B8F430 | [16, 19[ | m | AML | AML M2 | train | | | |
| F899139DF5E1059396431415E770C6DD | [3, 4[ | m | ALL | pre-B-ALL | test | | | |
| 93DB85ED909C13838FF95CCFA94CEBD9 | [0, 3[ | m | ALL | c-ALL | test | | | |
| 72B32A1F754BA1C09B3695E0CB6CDE7F | [4, 6[ | f | ALL | c-ALL | train | No cell classes annotated for this patient. | | |
| AD61AB143223EFBC24C7D2583BE69251 | [4, 6[ | m | ALL | c-ALL | test | | | |
| A0A080F42E6F13B3A2DF133F073095DD | [8, 11[ | m | AML | AML M4 | train | No cell classes annotated for this patient. | | |
| 1385974ED5904A438616FF7BDB3F7439 | [14, 16[ | f | AML | AML M1 | test | No cell classes annotated for this patient. | | |
| 2838023A778DFAECDC212708F721B788 | [3, 4[ | f | ALL | c-ALL, biphenotypic | test | No cell classes annotated for this patient. | | |
| 5FD0B37CD7DBBB00F97BA6CE92BF5ADD | [16, 19[ | m | ALL | pre-B-ALL | test | | | |
| 2B44928AE11FB9384C4CF38708677C48 | [14, 16[ | f | ALL | pre-B-ALL, co-expression | test | | | |
| D1F491A404D6854880943E5C3CD9CA25 | [11, 14[ | f | AML | AML M2 | train | | | |
| 4E732CED3463D06DE0CA9A15B6153677 | [14, 16[ | m | CML | Chronic Phase | train | | | |
| C45147DEE729311EF5B5C3003946C48F | [6, 8[ | m | ALL | c-ALL, co-expression | train | | | |
| D645920E395FEDAD7BBBED0ECA3FE2E0 | [11, 14[ | m | CML | Chronic Phase | train | | | |
| A3C65C2974270FD093EE8A9BF8AE7D0B | [0, 3[ | m | ALL | c-ALL | train | | | |
| C16A5320FA475530D9583C34FD356EF5 | [4, 6[ | m | CML | Chronic Phase | train | | | |
| 2A38A4A9316C49E5A833517C45D31070 | [0, 3[ | f | ALL | c-ALL | train | | | |
| 093F65E080A295F8076B1C5722A46AA2 | [6, 8[ | m | ALL | pre-B-ALL | validation | No cell classes annotated for this patient. | | |
| 7F6FFAA6BB0B408017B62254211691B5 | [0, 3[ | f | ALL | c-ALL, co-expression | train | | | |
| F7177163C833DFF4B38FC8D2872F1EC6 | [8, 11[ | m | CML | Chronic Phase | test | | | |
| 8F14E45FCEEA167A5A36DEDD4BEA2543 | [14, 16[ | m | CML | Blast Phase | validation | | | |
| D67D8AB4F4C10BF22AA353E27879133C | [8, 11[ | m | CML | Chronic Phase | train | | | No lab values available. |
| 26657D5FF9020D2ABEFE558796B99584 | [3, 4[ | m | ALL | c-ALL | test | | | |
| 8613985EC49EB8F757AE6439E879BB2A | [11, 14[ | m | ALL | intermediate cortical T-ALL | train | | | |
| 2723D092B63885E0D7C260CC007E8B9D | [6, 8[ | f | ALL | pro-B-ALL, co-expression | test | | | |
| 6C8349CC7260AE62E3B1396831A8398F | [14, 16[ | f | CML | Chronic Phase | train | | | |
| 3988C7F88EBCB58C6CE932B957B6F332 | [11, 14[ | m | AML | AML M5 | train | | | No lab values available. No clinical DCC available. |
| 4443AEE183B279F76A95C13C7F5BCA0D | [16, 19[ | m | CML | Blast Phase | test | | | |
| 6974CE5AC660610B44D9B9FED0FF9548 | [14, 16[ | m | ALL | c-ALL | validation | | | |
| 7F39F8317FBDB1988EF4C628EBA02591 | [0, 3[ | m | ALL | c-ALL | train | | | |
| C7E1249FFC03EB9DED908C236BD1996D | [4, 6[ | m | ALL | pre-B-ALL | train | | | |
| 45C48CCE2E2D7FBDEA1AFC51C7C6AD26 | [16, 19[ | f | CML | Chronic Phase | train | | | |
| 67C6A1E7CE56D3D6FA748AB6D9AF3FD7 | [14, 16[ | m | CML | Chronic Phase | validation | | | |
| EC8956637A99787BD197EACD77ACCE5E | [0, 3[ | f | ALL | pro-B-ALL | test | | | |
| D09BF41544A3365A46C9077EBB5E35C3 | [11, 14[ | m | ALL | c-ALL | test | | | |
| 6512BD43D9CAA6E02C990B0A82652DCA | [4, 6[ | m | CML | Blast Phase | train | | | |
| 43EC517D68B6EDD3015B3EDC9A11367B | [3, 4[ | m | ALL | c-ALL | test | | | |
| FC490CA45C00B1249BBE3554A4FDF6FB | [4, 6[ | f | ALL | intermediate cortical T-ALL | validation | No cell classes annotated for this patient. | | |
| 072B030BA126B2F4B2374F342BE9ED44 | [8, 11[ | m | ALL | pre-B-ALL | test | No cell classes annotated for this patient. | | |
| C0C7C76D30BD3DCAEFC96F40275BDC0A | [14, 16[ | f | CML | Chronic Phase | train | | | |
| 5EF059938BA799AAA845E1C2E8A762BD | [14, 16[ | m | ALL | c-ALL | train | | | |
| A6BBC91AE73DD21C0533F735470A9CD0 | [16, 19[ | f | CML | Chronic Phase | train | | | |
| C74D97B01EAE257E44AA9D5BADE97BAF | [16, 19[ | m | CML | Chronic Phase | validation | | | |
| 9F61408E3AFB633E50CDF1B20DE6F466 | [4, 6[ | m | ALL | c-ALL | train | No cell classes annotated for this patient. | | |
| FBD7939D674997CDB4692D34DE8633C4 | [3, 4[ | m | ALL | intermediate cortical T-ALL | train | | | |
| 9FC3D7152BA9336A670E36D0ED79BC43 | [16, 19[ | f | AML | AML M2 | train | | | |
| 9A1158154DFA42CADDBD0694A4E9BDC8 | [3, 4[ | f | ALL | c-ALL | train | No cell classes annotated for this patient. | | |
| 3DEF184AD8F4755FF269862EA77393DD | [14, 16[ | f | AML | AML M3 | test | | | |
| C7FF6173CE74C106F0871CEA7F91F152 | [6, 8[ | | ALL | Burkitt leukemia | validation | No cell classes annotated for this patient. | | No lab values available. |
| 3B80517688B93186611E64FE9268FCA7 | [8, 11[ | | ALL | Burkitt leukemia | train | No cell classes annotated for this patient. | | |
| 700049AE9F8CB7635BF38EA93D450C00 | [14, 16[ | | ALL | Burkitt leukemia | test | No cell classes annotated for this patient. | | |
| 0A09C8844BA8F0936C20BD791130D6B6 | [16, 19[ | m | AML | AML M1 | train | | | No lab values available. No clinical DCC available. |
| 2B24D495052A8CE66358EB576B8912C8 | [4, 6[ | f | AML | | train | | | No lab values available. No clinical DCC available. |
| 8D5E957F297893487BD98FA830FA6413 | [14, 16[ | f | AML | AML M3 | test | | | No lab values available. No clinical DCC available. |
| 47D1E990583C9C67424D369F3414728E | [11, 14[ | m | AML | AML M3 | train | | | No lab values available. No clinical DCC available. |
| F2217062E9A397A1DCA429E7D70BC6CA | [3, 4[ | m | AML | AML M5 | train | | | No lab values available. No clinical DCC available. |
| 7EF605FC8DBA5425D6965FBD4C8FBE1F | [4, 6[ | m | AML | | test | | | No lab values available. No clinical DCC available. |
| A8F15EDA80C50ADB0E71943ADC8015CF | [0, 3[ | m | AML | | validation | | | No lab values available. No clinical DCC available. |
| 37A749D808E46495A8DA1E5352D03CAE | [0, 3[ | f | AML | AML M7 | train | No cell classes annotated for this patient. | | No lab values available. No clinical DCC available. |
| B3E3E393C77E35A4A3F3CBD1E429B5DC | [16, 19[ | f | AML | AML M2 | train | No cell classes annotated for this patient. | | No lab values available. No clinical DCC available. |
| 2A79EA27C279E471F4D180B08D62B00A | [11, 14[ | f | AML | AML M1 | validation | No cell classes annotated for this patient. | | No lab values available. No clinical DCC available. |
| 1C9AC0159C94D8D0CBEDC973445AF2DA | [0, 3[ | f | AML | AML M4 | validation | No cell classes annotated for this patient. | | No lab values available. No clinical DCC available. |
| 6C4B761A28B734FE93831E3FB400CE87 | [8, 11[ | f | AML | AML M1 | train | No cell classes annotated for this patient. | | No lab values available. No clinical DCC available. |
| 06409663226AF2F3114485AA4E0A23B4 | [0, 3[ | f | AML | AML M5 | test | No cell classes annotated for this patient. | | No lab values available. No clinical DCC available. |
| 140F6969D5213FD0ECE03148E62E461E | [6, 8[ | m | AML | AML M1 | test | No cell classes annotated for this patient. | | No lab values available. No clinical DCC available. |
| B73CE398C39F506AF761D2277D853A92 | [0, 3[ | m | AML | AML M7 | validation | No cell classes annotated for this patient. | | No lab values available. No clinical DCC available. |
| BD4C9AB730F5513206B999EC0D90D1FB | [8, 11[ | f | AML | AML M2 | validation | No cell classes annotated for this patient. | | No lab values available. No clinical DCC available. |
| 82AA4B0AF34C2313A562076992E50AA3 | [14, 16[ | m | AML | AML M1 | train | No cell classes annotated for this patient. | | No lab values available. No clinical DCC available. |
| 0777D5C17D4066B82AB86DFF8A46AF6F | [11, 14[ | f | AML | AML M2 | test | No cell classes annotated for this patient. | | No lab values available. No clinical DCC available. |
| 9766527F2B5D3E95D4A733FCFB77BD7E | [3, 4[ | f | AML | AML M3 | validation | No cell classes annotated for this patient. | | No lab values available. No clinical DCC available. |
| 7E7757B1E12ABCB736AB9A754FFB617A | [0, 3[ | f | AML | AML M7 | train | No cell classes annotated for this patient. | | No lab values available. No clinical DCC available. |
| 5878A7AB84FB43402106C575658472FA | [0, 3[ | f | AML | AML M7 | validation | No cell classes annotated for this patient. | | No lab values available. No clinical DCC available. |
| 006F52E9102A8D3BE2FE5614F42BA989 | [11, 14[ | f | AML | AML M2 | train | No cell classes annotated for this patient. | | No lab values available. No clinical DCC available. |
| 3636638817772E42B59D74CFF571FBB3 | [3, 4[ | f | AML | AML M3 | train | No cell classes annotated for this patient. | | No lab values available. No clinical DCC available. |
| 149E9677A5989FD342AE44213DF68868 | [0, 3[ | f | AML | AML M7 | train | No cell classes annotated for this patient. | | No lab values available. No clinical DCC available. |
| A4A042CF4FD6BFB47701CBC8A1653ADA | [0, 3[ | f | AML | AML M4 | test | No cell classes annotated for this patient. | | No lab values available. No clinical DCC available. |
| BF8229696F7A3BB4700CFDDEF19FA23F | [0, 3[ | f | AML | AML M4 | train | No cell classes annotated for this patient. | | No lab values available. No clinical DCC available. |
| 82161242827B703E6ACF9C726942A1E4 | [0, 3[ | f | AML | AML M7 | train | No cell classes annotated for this patient. | | No lab values available. No clinical DCC available. |
| 8F85517967795EEEF66C225F7883BDCB | [16, 19[ | f | AML | AML M0 | test | No cell classes annotated for this patient. | | No lab values available. |
| FC221309746013AC554571FBD180E1C8 | [8, 11[ | m | AML | AML M0 | train | No cell classes annotated for this patient. | | No lab values available. |
| 31FEFC0E570CB3860F2A6D4B38C6490D | [16, 19[ | m | AML | AML M1 | validation | No cell classes annotated for this patient. | | No lab values available. No clinical DCC available. |
| 9DCB88E0137649590B755372B040AFAD | [14, 16[ | m | AML | AML M1 | train | No cell classes annotated for this patient. | | No lab values available. No clinical DCC available. |
| 0AA1883C6411F7873CB83DACB17B0AFC | [14, 16[ | f | AML | AML M1 | test | No cell classes annotated for this patient. | | No lab values available. |
| A597E50502F5FF68E3E25B9114205D4A | [14, 16[ | | AML | AML M1 | train | No cell classes annotated for this patient. | | No lab values available. No clinical DCC available. |
| 0336DCBAB05B9D5AD24F4333C7658A0E | [4, 6[ | | AML | AML M1 | train | No cell classes annotated for this patient. | | No lab values available. |
| BD686FD640BE98EFAAE0091FA301E613 | [14, 16[ | f | AML | AML M1 | train | No cell classes annotated for this patient. | | No lab values available. |
| 58A2FC6ED39FD083F55D4182BF88826D | [14, 16[ | f | AML | AML M1 | validation | No cell classes annotated for this patient. | | No lab values available. |
| 084B6FBB10729ED4DA8C3D3F5A3AE7C9 | [11, 14[ | m | AML | AML M1 | test | No cell classes annotated for this patient. | | No lab values available. |
| 84D9EE44E457DDEF7F2C4F25DC8FA865 | [3, 4[ | f | AML | AML M1 | train | No cell classes annotated for this patient. | | No lab values available. |
| 3644A684F98EA8FE223C713B77189A77 | [14, 16[ | f | AML | AML M1 | train | No cell classes annotated for this patient. | | No lab values available. |
| E2C0BE24560D78C5E599C2A9C9D0BBD2 | [8, 11[ | m | AML | AML M2 | validation | No cell classes annotated for this patient. | | No lab values available. |
| 274AD4786C3ABCA69FA097B85867D9A4 | [6, 8[ | m | AML | AML M2 | train | No cell classes annotated for this patient. | | No lab values available. |
| EAE27D77CA20DB309E056E3D2DCD7D69 | [8, 11[ | f | AML | AML M2 | train | No cell classes annotated for this patient. | | No lab values available. No clinical DCC available. |
| 69ADC1E107F7F7D035D7BAF04342E1CA | [6, 8[ | f | AML | AML M2 | validation | No cell classes annotated for this patient. | | No lab values available. |
| 091D584FCED301B442654DD8C23B3FC9 | [6, 8[ | m | AML | AML M2 | test | No cell classes annotated for this patient. | | No lab values available. |
| B1D10E7BAFA4421218A51B1E1F1B0BA2 | [8, 11[ | f | AML | AML M2 | train | No cell classes annotated for this patient. | | No lab values available. |
| 979D472A84804B9F647BC185A877A8B5 | [11, 14[ | f | AML | AML M2 | test | No cell classes annotated for this patient. | | No lab values available. |
| CA46C1B9512A7A8315FA3C5A946E8265 | [3, 4[ | f | AML | AML M2 | train | No cell classes annotated for this patient. | | No lab values available. |
| 3B8A614226A953A8CD9526FCA6FE9BA5 | [16, 19[ | m | AML | AML M2 | train | No cell classes annotated for this patient. | | No lab values available. |
| 63DC7ED1010D3C3B8269FAF0BA7491D4 | [16, 19[ | f | AML | AML M2 | validation | No cell classes annotated for this patient. | | No lab values available. |
| E96ED478DAB8595A7DBDA4CBCBEE168F | [4, 6[ | f | AML | AML M2 | train | No cell classes annotated for this patient. | | No lab values available. |
| 6F3EF77AC0E3619E98159E9B6FEBF557 | [11, 14[ | m | AML | AML M2 | train | No cell classes annotated for this patient. | | No lab values available. |
| D1C38A09ACC34845C6BE3A127A5AACAF | [8, 11[ | f | AML | AML M3 | train | No cell classes annotated for this patient. | | No lab values available. |
| EC8CE6ABB3E952A85B8551BA726A1227 | [14, 16[ | m | AML | AML M3 | train | No cell classes annotated for this patient. | | No lab values available. |
| 060AD92489947D410D897474079C1477 | [14, 16[ | m | AML | AML M3 | validation | No cell classes annotated for this patient. | | No lab values available. |
| C0E190D8267E36708F955D7AB048990D | [11, 14[ | f | AML | AML M3 | test | No cell classes annotated for this patient. | | No lab values available. |
| 9B04D152845EC0A378394003C96DA594 | [14, 16[ | f | AML | AML M4 | train | No cell classes annotated for this patient. | | No lab values available. |
| 74DB120F0A8E5646EF5A30154E9F6DEB | [3, 4[ | m | AML | AML M4 | test | No cell classes annotated for this patient. | | No lab values available. |
| E56954B4F6347E897F954495EAB16A88 | [0, 3[ | f | AML | AML M4Eo | train | No cell classes annotated for this patient. | | No lab values available. |
| EDA80A3D5B344BC40F3BC04F65B7A357 | [0, 3[ | m | AML | AML M4Eo | train | No cell classes annotated for this patient. | | No lab values available. |
| 8F121CE07D74717E0B1F21D122E04521 | [11, 14[ | m | AML | AML M4Eo | validation | No cell classes annotated for this patient. | | No lab values available. |
| 06138BC5AF6023646EDE0E1F7C1EAC75 | [11, 14[ | f | AML | AML M4Eo | test | No cell classes annotated for this patient. | | No lab values available. |
| 39059724F73A9969845DFE4146C5660E | [6, 8[ | m | AML | AML M4Eo | train | No cell classes annotated for this patient. | | No lab values available. |
| 7F100B7B36092FB9B06DFB4FAC360931 | [16, 19[ | | AML | AML M4Eo | train | No cell classes annotated for this patient. | | No lab values available. |
| 7A614FD06C325499F1680B9896BEEDEB | [16, 19[ | m | AML | AML M4Eo | train | No cell classes annotated for this patient. | | No lab values available. |
| 335F5352088D7D9BF74191E006D8E24C | [8, 11[ | f | AML | AML M5 | train | No cell classes annotated for this patient. | | No lab values available. |
| E4A6222CDB5B34375400904F03D8E6A5 | [0, 3[ | f | AML | AML M5 | train | No cell classes annotated for this patient. | | No lab values available. |
| CB70AB375662576BD1AC5AAF16B3FCA4 | [16, 19[ | f | AML | AML M5 | validation | No cell classes annotated for this patient. | | No lab values available. |
| 38DB3AED920CF82AB059BFCCBD02BE6A | [3, 4[ | f | AML | AML M5 | test | No cell classes annotated for this patient. | | No lab values available. |
| 621BF66DDB7C962AA0D22AC97D69B793 | [0, 3[ | m | AML | AML M5 | train | No cell classes annotated for this patient. | | No lab values available. |
| C24CD76E1CE41366A4BBE8A49B02A028 | [0, 3[ | m | AML | AML M6 | test | No cell classes annotated for this patient. | | No lab values available. |
| 03C6B06952C750899BB03D998E631860 | [0, 3[ | m | AML | AML M6 | train | No cell classes annotated for this patient. | | No lab values available. |
| C52F1BD66CC19D05628BD8BF27AF3AD6 | [0, 3[ | m | AML | AML M7 | test | No cell classes annotated for this patient. | | No lab values available. |
| B1A59B315FC9A3002CE38BBE070EC3F5 | [4, 6[ | f | AML | AML M7 | train | No cell classes annotated for this patient. | | No lab values available. |
| D96409BF894217686BA124D7356686C9 | [0, 3[ | m | AML | AML M7 | validation | No cell classes annotated for this patient. | | No lab values available. |
| 36660E59856B4DE58A219BCF4E27EBA3 | [0, 3[ | m | AML | AML M7 | train | No cell classes annotated for this patient. | | No lab values available. |
| 2EAB84C785399B2DCDD75FD40917C17C | [4, 6[ | f | ALL | pre-B-ALL, co-expression | train | No cell classes annotated for this patient. | No cell boxes annotated for this patient. | No lab values available. No clinical DCC available. |
| 398499203B70C029E7F1BEFF75C0DEA1 | [0, 3[ | m | ALL | pre-B-ALL, co-expression | train | No cell classes annotated for this patient. | No cell boxes annotated for this patient. | No lab values available. No clinical DCC available. |
| BA6484A665B9A691A37F57872C643E04 | [14, 16[ | m | ALL | pro-B-ALL, co-expression | train | No cell classes annotated for this patient. | No cell boxes annotated for this patient. | No lab values available. No clinical DCC available. |
| F768F6C7948E6A750060AF15FB73BF6C | [4, 6[ | m | ALL | c-ALL, co-expression | test | No cell classes annotated for this patient. | No cell boxes annotated for this patient. | No lab values available. No clinical DCC available. |
| 5C820C4E9439DC590E3192DBB774255E | [0, 3[ | f | ALL | pre-B-ALL, co-expression | train | No cell classes annotated for this patient. | No cell boxes annotated for this patient. | No lab values available. No clinical DCC available. |
| F0549529AF3E0E57F00F30CF1E1FC994 | [8, 11[ | f | ALL | c-ALL, co-expression | test | No cell classes annotated for this patient. | No cell boxes annotated for this patient. | No lab values available. No clinical DCC available. |
| F7405A633A5F7482725CC94B3563A5B6 | [4, 6[ | f | ALL | c-ALL, co-expression | validation | No cell classes annotated for this patient. | No cell boxes annotated for this patient. | No lab values available. No clinical DCC available. |
| 399C0FDBA7E789C6736AC660D23B4BB6 | [4, 6[ | m | ALL | c-ALL, co-expression | validation | No cell classes annotated for this patient. | No cell boxes annotated for this patient. | No lab values available. No clinical DCC available. |
| 3421557EEC62C83EEA2C83C08A79CDB9 | [0, 3[ | f | ALL | c-ALL, co-expression | train | No cell classes annotated for this patient. | No cell boxes annotated for this patient. | No lab values available. No clinical DCC available. |
| BCDCEF5EF4CF9871F0CB6BFF04753754 | [0, 3[ | f | ALL | pre-B-ALL, co-expression | test | No cell classes annotated for this patient. | No cell boxes annotated for this patient. | No lab values available. No clinical DCC available. |
| 1A00B3B296CD48BA5D1BD76B18C48346 | [0, 3[ | m | ALL | pro-B-ALL, co-expression | validation | No cell classes annotated for this patient. | No cell boxes annotated for this patient. | No lab values available. No clinical DCC available. |
| 12ED127BACFF11D62A255BE81C5E2310 | [4, 6[ | m | ALL | c-ALL, co-expression | train | No cell classes annotated for this patient. | No cell boxes annotated for this patient. | No lab values available. No clinical DCC available. |
| 47397D3C4DB6F903852E87660CB96E3D | [4, 6[ | m | ALL | c-ALL, co-expression | train | No cell classes annotated for this patient. | No cell boxes annotated for this patient. | No lab values available. No clinical DCC available. |
| BB8CF22800F0D97AB7EC98F1213B4B4E | [6, 8[ | f | ALL | pre-B-ALL, co-expression | validation | No cell classes annotated for this patient. | No cell boxes annotated for this patient. | No lab values available. No clinical DCC available. |
| 4F584FE84E3D2EDE914508A873D75855 | [3, 4[ | f | ALL | c-ALL, biphenotypic | train | No cell classes annotated for this patient. | No cell boxes annotated for this patient. | No lab values available. No clinical DCC available. |
| 357AAD4942E92FCC103D9233B6739697 | [0, 3[ | f | ALL | pro-B-ALL | train | No cell classes annotated for this patient. | No cell boxes annotated for this patient. | No lab values available. No clinical DCC available. |
| B4A67623E7A47A4406DECAC8E1A1DD6F | [6, 8[ | f | ALL | c-ALL, biphenotypic | validation | No cell classes annotated for this patient. | No cell boxes annotated for this patient. | No lab values available. No clinical DCC available. |
| 002BD524C907A2A5121D3ADAED05D442 | [11, 14[ | f | ALL | intermediate cortical T-ALL | train | No cell classes annotated for this patient. | No cell boxes annotated for this patient. | No lab values available. No clinical DCC available. |
| A61D1A964CA038E6FB5416A8ECF0C76D | [8, 11[ | f | ALL | intermediate cortical T-ALL | train | No cell classes annotated for this patient. | No cell boxes annotated for this patient. | No lab values available. No clinical DCC available. |
| EAA51AE97DBF78D298CB4BA55761AB8B | [11, 14[ | m | ALL | pro-T-ALL | train | No cell classes annotated for this patient. | No cell boxes annotated for this patient. | No lab values available. No clinical DCC available. |
| 93ED4838BE11F21D2E5A8A03B36CD6BD | [8, 11[ | m | ALL | c-ALL, co-expression | train | No cell classes annotated for this patient. | No cell boxes annotated for this patient. | No lab values available. No clinical DCC available. |
| 61BEED9E2C7DBF9CF3F2DBA9FFD26B2A | [6, 8[ | m | ALL | c-ALL, co-expression | train | No cell classes annotated for this patient. | No cell boxes annotated for this patient. | No lab values available. No clinical DCC available. |
| C24D731AC18944F0AFF1DBCCF40ED102 | [4, 6[ | m | ALL | intermediate cortical T-ALL | train | No cell classes annotated for this patient. | No cell boxes annotated for this patient. | No lab values available. No clinical DCC available. |
| ACC68D832D90A3AAA9E2A38A529C62EB | [11, 14[ | f | ALL | c-ALL, co-expression | train | No cell classes annotated for this patient. | No cell boxes annotated for this patient. | No lab values available. No clinical DCC available. |
| 9461E2E0784D432D5FD84528FDBC3BBE | [6, 8[ | m | ALL | intermediate cortical T-ALL | train | No cell classes annotated for this patient. | No cell boxes annotated for this patient. | No lab values available. No clinical DCC available. |
| D25E9132CF24B8A7A2164885101E62CB | [16, 19[ | f | ALL | pro-T-ALL | validation | No cell classes annotated for this patient. | No cell boxes annotated for this patient. | No lab values available. No clinical DCC available. |
| 8963A6178314866F22F870E84026D429 | [16, 19[ | m | ALL | intermediate cortical T-ALL | test | No cell classes annotated for this patient. | No cell boxes annotated for this patient. | No lab values available. No clinical DCC available. |
| 12F54ACB6F92C80EAC9609D202A98325 | [4, 6[ | f | ALL | intermediate cortical T-ALL | test | No cell classes annotated for this patient. | No cell boxes annotated for this patient. | No lab values available. No clinical DCC available. |
| 75612D01831B0B8C48530A878109A865 | [6, 8[ | m | ALL | intermediate cortical T-ALL | validation | No cell classes annotated for this patient. | No cell boxes annotated for this patient. | No lab values available. No clinical DCC available. |
| 06FEE55734DEFE2EF2A60DEFFFD994FE | [14, 16[ | m | ALL | intermediate cortical T-ALL | validation | No cell classes annotated for this patient. | No cell boxes annotated for this patient. | No lab values available. No clinical DCC available. |
| C912541E67CD22445ED91775B8E89A7D | [14, 16[ | f | ALL | Burkitt leukemia | train | No cell classes annotated for this patient. | No cell boxes annotated for this patient. | No lab values available. No clinical DCC available. |
| 87D5BE18C3F5EA9721B8B343120CD71C | [11, 14[ | m | ALL | Burkitt leukemia | train | No cell classes annotated for this patient. | No cell boxes annotated for this patient. | No lab values available. No clinical DCC available. |
| 9A15F38C5291BA675D2E6DDF530BA6D9 | [8, 11[ | m | ALL | Burkitt leukemia | validation | No cell classes annotated for this patient. | No cell boxes annotated for this patient. | No lab values available. No clinical DCC available. |
| 623AE0B81861352D483A3CAE6CD34DDB | [8, 11[ | m | ALL | Burkitt leukemia | test | No cell classes annotated for this patient. | No cell boxes annotated for this patient. | No lab values available. No clinical DCC available. |
| FFD52F3C7E12435A724A8F30FDDADD9C | [0, 3[ | f | AML | AML M0 | validation | No cell classes annotated for this patient. | No cell boxes annotated for this patient. | No lab values available. No clinical DCC available. |

**Supplemental Table 3**: Clinical data and stratification of patients, as well as information about exclusion of specific patients from certain tasks.

| Cell Class | F1 | Recall | Precision |
|---|---|---|---|
| Myeloid Precursor Cell | 0.409 (0.186-0.649) | 0.409 (0.208-0.667) | 0.409 (0.149-0.697) |
| Myelocytic Blast | 0.490 (0.154-0.694) | 0.489 (0.129-0.764) | 0.492 (0.155-0.674) |
| Promyelocyte | 0.799 (0.765-0.843) | 0.850 (0.795-0.911) | 0.754 (0.694-0.802) |
| Neutrophilic Myelocyte | 0.543 (0.458-0.628) | 0.543 (0.430-0.644) | 0.543 (0.448-0.626) |
| Neutrophilic Metamyelocyte | 0.588 (0.519-0.661) | 0.589 (0.488-0.673) | 0.586 (0.507-0.662) |
| Neutrophilic Band | 0.724 (0.678-0.770) | 0.745 (0.679-0.810) | 0.704 (0.656-0.753) |
| Segmented Neutrophil | 0.907 (0.885-0.927) | 0.882 (0.850-0.916) | 0.933 (0.897-0.959) |
| Eosinophilic Myelocyte | 0.731 (0.566-0.839) | 0.850 (0.629-0.977) | 0.642 (0.481-0.763) |
| Eosinophilic Metamyelocyte | 0.194 (0.000-0.485) | 0.158 (0.000-0.473) | 0.250 (0.000-0.600) |
| Eosinophilic Band | 0.286 (0.080-0.539) | 0.267 (0.063-0.636) | 0.308 (0.091-0.566) |
| Segmented Eosinophil | 0.720 (0.613-0.875) | 0.726 (0.519-0.868) | 0.714 (0.549-0.897) |
| Immature Basophil | 0.250 (0.000-0.646) | 0.176 (0.000-0.500) | 0.429 (0.000-1.000) |
| Segmented Basophil | 0.615 (0.420-0.799) | 0.600 (0.370-0.861) | 0.632 (0.348-0.811) |
| Immature Monocyte | 0.261 (0.000-0.539) | 0.176 (0.000-0.400) | 0.500 (0.000-1.000) |
| Monocytic Blast | 0.358 (0.138-0.630) | 0.370 (0.132-0.639) | 0.347 (0.118-0.654) |
| Monocyte | 0.241 (0.089-0.471) | 0.189 (0.060-0.403) | 0.333 (0.130-0.660) |
| Immature Lymphocyte | 0.000 (0.000-0.000) | 0.000 (0.000-0.000) | 0.000 (0.000-0.000) |
| Lymphocytic Blast | 0.886 (0.853-0.922) | 0.924 (0.887-0.962) | 0.851 (0.803-0.894) |
| Lymphocyte | 0.450 (0.272-0.604) | 0.417 (0.248-0.591) | 0.490 (0.266-0.688) |
| Proerythroblast or Basophilic Erythroblast | 0.439 (0.218-0.683) | 0.643 (0.300-0.928) | 0.333 (0.143-0.610) |
| Polychromatic Erythroblast | 0.648 (0.506-0.821) | 0.719 (0.531-0.899) | 0.590 (0.439-0.800) |
| Orthochromatic Erythroblast | 0.764 (0.667-0.850) | 0.714 (0.600-0.821) | 0.821 (0.693-0.917) |
| Megakaryocyte | 0.945 (0.888-0.986) | 0.992 (0.952-1.000) | 0.903 (0.806-0.980) |
| Thrombocyte | 0.931 (0.874-0.970) | 0.899 (0.830-0.954) | 0.965 (0.881-0.994) |
| Giant Platelet | 0.129 (0.000-0.387) | 0.143 (0.000-0.441) | 0.118 (0.000-0.480) |
| Neutrophil Extracellular Trap | 0.876 (0.818-0.939) | 0.878 (0.794-0.949) | 0.874 (0.808-0.942) |
| Pseudo Gaucher Cell | 0.667 (0.308-1.000) | 0.800 (0.167-1.000) | 0.571 (0.273-1.000) |
| Mitosis | 0.692 (0.430-0.901) | 0.692 (0.385-0.923) | 0.692 (0.310-1.000) |
| Spicule | 0.938 (0.880-0.970) | 0.953 (0.843-1.000) | 0.924 (0.833-0.978) |
| Other Cell | 0.794 (0.630-0.891) | 0.718 (0.559-0.836) | 0.889 (0.657-0.987) |
| Smudge Cell | 0.676 (0.576-0.756) | 0.722 (0.638-0.802) | 0.635 (0.522-0.734) |
| Artifact | 0.557 (0.466-0.657) | 0.582 (0.470-0.693) | 0.534 (0.422-0.643) |
| Not Identifiable | 0.531 (0.465-0.597) | 0.474 (0.403-0.549) | 0.602 (0.510-0.671) |

**Supplemental Table 4**: Precision, recall, and F1-scores including 95% confidence intervals (BCa bootstrap) of the cell classifier per class.